\documentclass[12pt]{article}

\usepackage{scicite}
\usepackage{booktabs}
\usepackage{times}
\usepackage{hyperref} 
\usepackage{graphicx}
\usepackage{amsmath}
\usepackage[ruled,vlined,noend,linesnumbered]{algorithm2e}
\usepackage{xcolor}

\newcommand{\supplementarysection}{%
  \setcounter{figure}{0}
  \let\oldthefigure\thefigure
  \renewcommand{\thefigure}{S\oldthefigure}
  \section*{Supplementary Materials}
}

\newenvironment{sciabstract}{%
\begin{quote} \bf}
{\end{quote}}

\title{Few-shot Scooping Under Domain Shift via Simulated Maximal Deployment Gaps}

\author
{Yifan Zhu$^{1\ast\dag}$, Pranay Thangeda$^{2\dag}$, Erica L Tevere$^{3}$, Ashish Goel$^{3}$, Erik Kramer$^{4}$, \\Hari D Nayar$^{3}$, Melkior Ornik$^{2}$, Kris Hauser$^{5}$\\
\\
\normalsize{$^{1}$Department of Mechanical Engineering and Material Science, Yale University}\\
\normalsize{New Haven, CT, USA}\\
\normalsize{$^{2}$Department of Aerospace Engineering, University of Illinois Urbana-Champaign}\\
\normalsize{Champaign, IL, USA}\\
\normalsize{$^{3}$Jet Propulsion Laboratory, California Institute of Technology}\\
\normalsize{Pasadena, CA, USA}\\
\normalsize{$^{4}$Department of Mechanical and Aerospace Engineering, University of California, Los Angeles}\\
\normalsize{Los Angeles, CA, USA}\\
\normalsize{$^{5}$Department of Computer Science, University of Illinois Urbana-Champaign}\\
\normalsize{Champaign, IL, USA}\\
\\
\normalsize{$^\dag$These authors contributed equally to this work.}
\\
\normalsize{$^\ast$To whom correspondence should be addressed; E-mail:  yifan.zhu@yale.edu}
}

\date{}

\begin{document} 


\baselineskip24pt


\maketitle


\begin{sciabstract}
Autonomous lander missions on extraterrestrial bodies need to sample granular materials while coping with domain shifts, even when sampling strategies are extensively tuned on Earth. To tackle this challenge, this paper studies the few-shot scooping problem and proposes a vision-based adaptive scooping strategy that uses the deep kernel Gaussian process method trained with a novel meta-training strategy to learn online from very limited experience on out-of-distribution target terrains. Our Deep Kernel Calibration with Maximal Deployment Gaps (kCMD) strategy explicitly trains a deep kernel model to adapt to large domain shifts by creating simulated maximal deployment gaps from an offline training dataset and training models to overcome these deployment gaps during training. Employed in a Bayesian Optimization sequential decision-making framework, the proposed method allows the robot to perform high-quality scooping actions on out-of-distribution terrains after a few attempts, significantly outperforming non-adaptive methods proposed in the excavation literature as well as other state-of-the-art meta-learning methods. The proposed method also demonstrates zero-shot transfer capability, successfully adapting to the NASA OWLAT platform, which serves as a state-of-the-art simulator for potential future planetary missions. These results demonstrate the potential of training deep models with simulated deployment gaps for more generalizable meta-learning in high-capacity models. Furthermore, they highlight the promise of our method in autonomous lander sampling missions by enabling landers to overcome the deployment gap between Earth and extraterrestrial bodies.
\end{sciabstract}

\paragraph{One-Sentence Summary:} An adaptive scooping robot that learns to sample unknown terrains in a few attempts for extraterrestrial exploration.

\section*{Introduction}
Terrain sampling with landers and rovers during extraterrestrial scientific explorations is typically done with humans in the loop where a team of experts would carefully teleoperate the robot from the earth~\cite{von2022surface}, which is prohibitively slow, suffers from long delays, and can be interrupted for long durations. Meanwhile, autonomous sampling has the potential to increase efficiency drastically but faces daunting challenges, including large uncertainties in terrain material properties and composition, restrictions in onboard computation, and a limited sampling capacity. 
\begin{figure}[ht!]
\centering
    \includegraphics[trim=0cm 4.8cm 15.5cm 0cm,clip,width=0.9\linewidth]{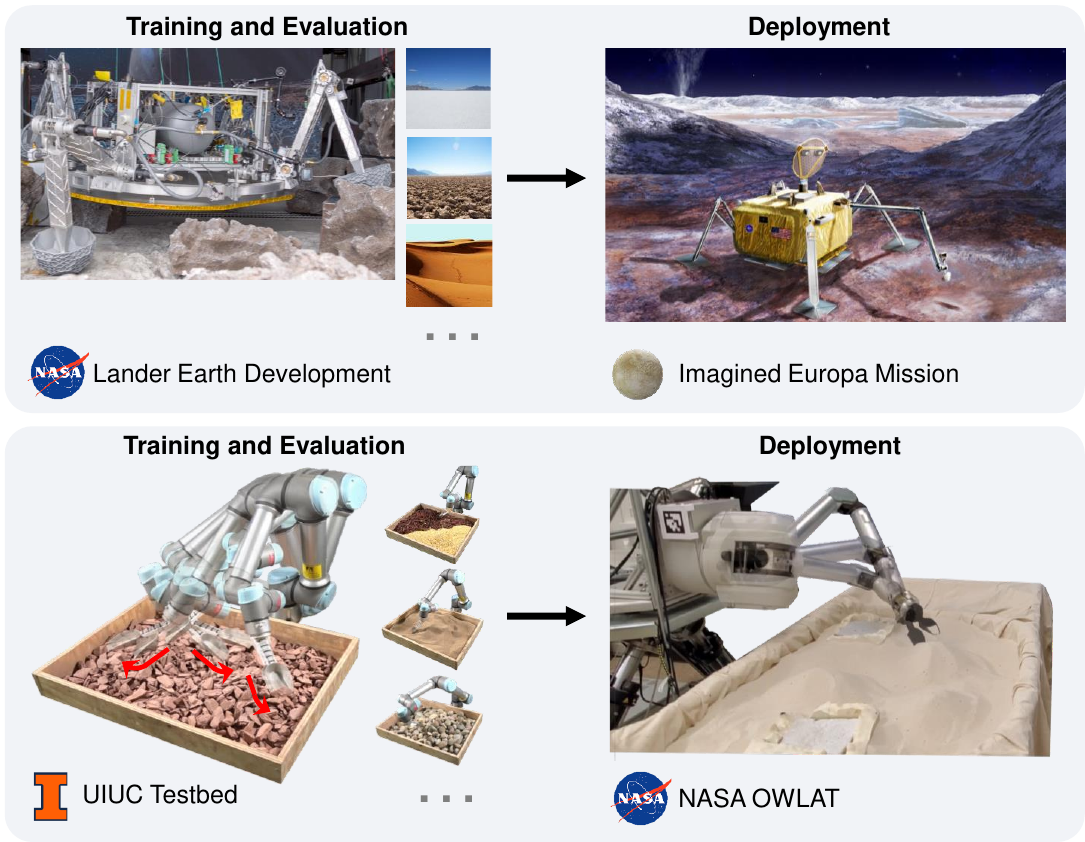}
    \caption{\textbf{Concept illustration.} A lander whose sampling policy is trained and tuned on Earth may degrade or even completely fail when deployed on an extraterrestrial planet with drastically different terrain properties (pictures by courtesy of NASA). To overcome such a challenge, our scooping policy is trained to be adaptive to novel terrains on a large offline dataset collected on the UIUC testbed and evaluated on novel terrains in the same testbed. The policy is then deployed, without retraining, on the NASA OWLAT platform with its novel terrains, where the policy quickly adapts, achieving high scooping volumes in just a few attempts.}
    \label{fig:platforms}
\end{figure}
As illustrated in Fig.~\ref{fig:platforms}, our work is inspired by the proposed NASA missions to send autonomous landers to Europa and Enceladus for collecting and analyzing terrain samples, exploring whether these bodies exhibit conditions that could support extraterrestrial life~\cite{NASA2017}. In this mission, not only is autonomous sampling desirable, but it is necessary: the mission is designed to last only 20 to 40 days due to the adversarial conditions on Europa and Enceladus that would greatly limit teleoperated operations. However, while engineers can implement autonomous sampling by tuning scripted sampling policies on terrain simulants on Earth, these policies will inevitably face a {\em deployment gap} when operating on an extraterrestrial body when the terrain properties are significantly different from the materials the policies were tested on. In such cases, the lander could degrade or completely fail to collect samples. In order to address these challenges, it is essential for a robot to quickly adjust its behavior based on a few failed attempts in unknown environments. To this end, we have developed a novel approach for a robot sampling system that learns from raw vision data and adapts quickly to drastically different scenarios.

Recently, deep-learning-based approaches have demonstrated the potential of allowing robots to solve vision-based problems~\cite{DRL-survey,DL-robotics-survey}, and learn extremely complex mappings with high-capacity neural networks. However, the performance of neural networks can deteriorate significantly when there is a deployment gap~\cite{candela2009dataset}, and adapting high-capacity networks on sparse online data tends to result in overfitting~\cite{hospedales2021meta}. The future of having robots deployed ubiquitously in the real world, where deployment gaps are frequent and inevitable, necessitates strategies that train high-capacity models that are adaptive to large domain gaps. One promising approach is few-shot meta-learning~\cite{hospedales2021meta}, which involves extracting useful information from an offline dataset consisting of multiple different but related tasks such that the model adapts to a novel task quickly. However, it is challenging to apply current few-shot meta-learning
methods that are mainly designed for computer vision tasks to the robot manipulation domain, and as shown in the results, current state-of-the-art
methods struggle with terrain sampling. Our solution is a few-shot meta-learning approach that {\em explicitly trains deep models to overcome large domain gaps}, where the training procedure creates large {\em simulated deployment gaps} from an offline training dataset and forces models during training to learn to overcome these deployment gaps. 
\begin{figure}[h!]
\centering
    \includegraphics[trim=0cm 9.5cm 11.8cm 0cm,clip,width=1\linewidth]{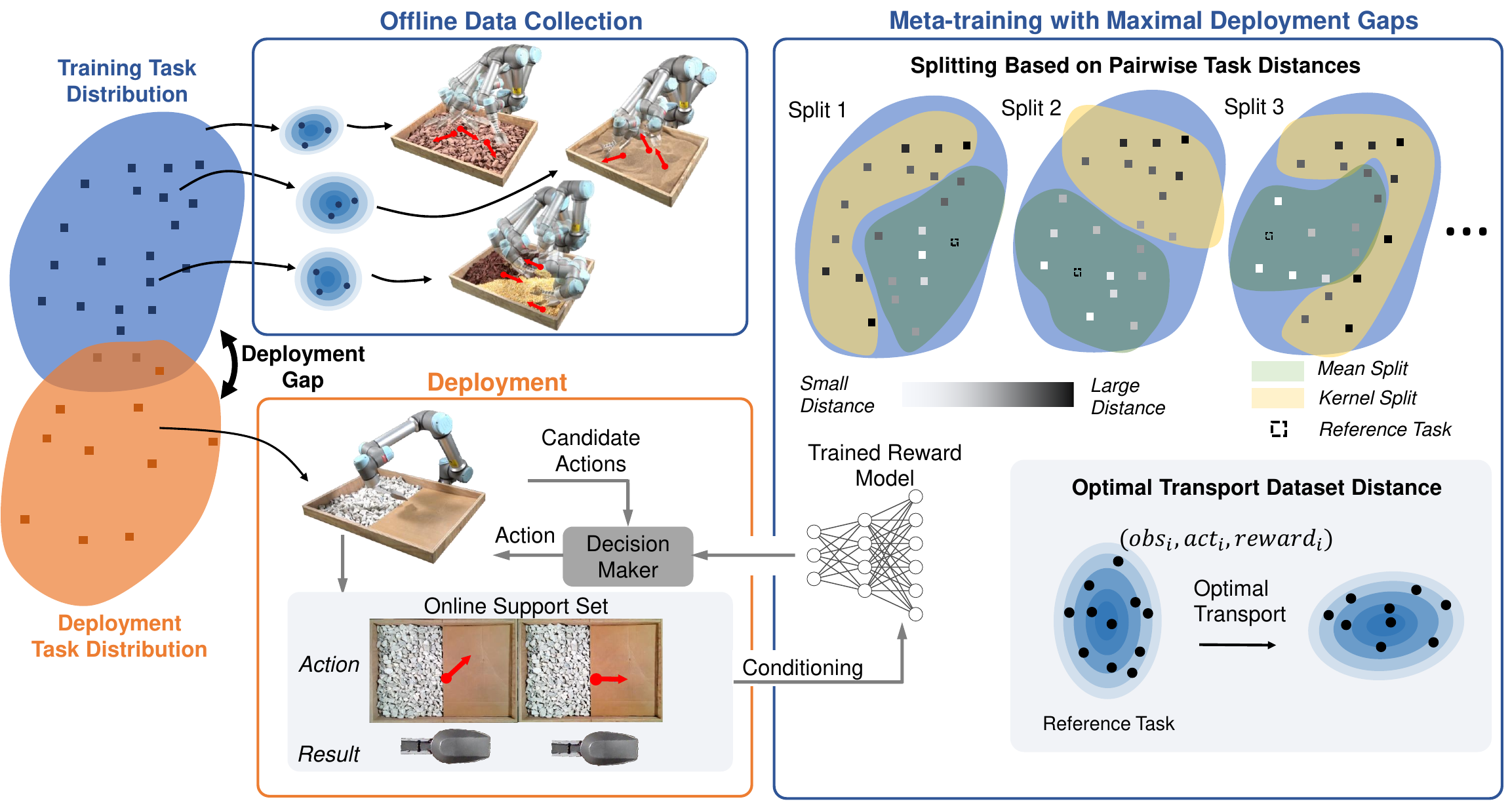}
    \caption{\textbf{Method overview.} Our proposed deep kernel model is trained on a diverse offline database with kCMD, which repeatedly splits the training set into mean-training and kernel-training and learns kernel parameters to minimize the residuals from the mean models. Mean- and kernel-training splits are achieved by randomly selecting a reference task, calculating the pairwise task distance between it and every other task (the gray-scale color represents the distance), and splitting based on the median distance. The task distance is the optimal transport~\cite{alvarez2020geometric} between the tasks based on each task's data samples. In deployment, the decision-maker uses the trained model and adapts it to the data acquired online (support set). }
    \label{fig:method}
\end{figure}

In this work, we study the few-shot scooping problem, in which the goal is to collect high-volume samples from a novel target terrain with a limited budget of attempts, where different terrain compositions and shapes require very different scooping strategies. During scooping, our model takes an RGB-D image of the terrain and parameters of a scooping action as input and predicts the mean and variance of the scooped volume. The model is a deep kernel Gaussian process (GP) method~\cite{wilson2016DKL} that employs a deep mean function and a deep kernel. A deep kernel transforms the input by a neural network before inputting it to a GP kernel, allowing GP to work with high-dimensional inputs. Compared to parametric models, the non-parametric deep kernel GP method has demonstrated good performance in few-shot learning tasks, exhibiting robustness against overfitting when dealing with limited data~\cite{Patacchiola2020DKT}. Summarized in Fig.~\ref{fig:method}, our Deep Kernel Calibration with Maximal Deployment Gaps (kCMD) method explicitly creates simulated maximal deployment gaps for the deep kernel to overcome by repeatedly splitting the training set into mean-training and kernel-training and learns kernel parameters to minimize loss over the residuals from the mean models. The splitting process maximizes the domain gaps between the mean-training and kernel-training sets by maximizing the optimal transport (OT)~\cite{torres2021survey} between the tasks in the two splits. This procedure trains the kernel on residuals that are more representative of the residuals seen in out-of-distribution (OOD) tasks, calibrating the kernel to OOD target tasks for fast adaption. 

Our model is trained on an offline training dataset\footnote{https://tinyurl.com/scooping-data} consisting of 5,100 scoops on a variety of terrains with different compositions and materials on a university (UIUC) testbed, shown in Fig.~\ref{fig:platforms}. For decision-making, we employ a Bayesian optimization (BO) framework. This framework selects actions by maximizing an acquisition function, balancing between scoop volume prediction and associated uncertainty. Our experiments first evaluate the proposed method on the UIUC testbed using OOD terrains that have drastically different appearances and/or material properties than the training terrains. kCMD allows the robot to achieve high-volume scooping actions on out-of-distribution terrains in a few attempts, outperforming state-of-the-art meta-learning methods. Moreover, it significantly outperforms non-adaptive methods such as those proposed in the granular material manipulation literature~\cite{Lu2021Excavation,yang2021optimization}. The effectiveness of our approach stems from the meta-training procedure, which forces the model to learn to adapt to large deployment gaps. Our results suggest that training models by leveraging large simulated deployment gaps is an effective approach to building adaptive high-capacity models. In addition, while kCMD is instantiated for the non-parametric deep kernel GP model, the idea of maximal deployment gaps can be adapted to improve the performance of parametric models: for instance, several works for parametric models have also proposed to split the training data to improve generalization~\cite{li2018learning, balaji2018metareg} but the splitting is done randomly.

Furthermore, we deploy our model as-is on the NASA JPL Ocean Worlds Lander Autonomy Testbed (OWLAT)~\cite{Nayar2021}, which serves as a state-of-the-art platform for simulating various potential future planetary missions. Our method quickly adapts to novel terrains on OWLAT out-of-the-box and also outperforms non-adaptive scooping methods. These results show the potential for employing our method for autonomous lander sampling missions, where the adaptability of the trained models allows the lander to overcome the deployment gaps between Earth and extraterrestrial bodies. We also would like to highlight that a paradigm shift for the robotics community is necessary toward a future of wide deployment of robot systems in the real world, where we should be explicitly building and evaluating practical robot systems for how well they can adapt to deployment gaps instead of testing on known environments in controlled lab settings.

To summarize, the contributions of our paper include:
\begin{itemize}
    \item A novel meta-training procedure, kCMD, for training deep kernels to be adaptive by training the kernels to overcome maximal simulated deployment gaps during training.
    \item A vision-based scooping system that is able to quickly acquire high-volume samples on novel terrains, along with extensive physical experiments on the UIUC testbed and direct transfer experiments on the OWLAT testbed. The systems were introduced in our conference publications~\cite{thangeda2024learning, zhu2023CoDeGa}.
\end{itemize}

\subsection*{Background and Related Work}
The problem of robotic scooping is highly related to the broader field of granular material manipulation, which encompasses a diverse range of real-world robotic applications from food preparation to construction and outdoor navigation. While recent research explored various aspects of granular manipulation—including pushing~\cite{Suh2021Pile}, grasping~\cite{Takahashi2021GraspingFoods}, untangling~\cite{Ray2020Untangling}, and locomotion~\cite{Shrivastava2020,Karsai2022}—our focus aligns most closely with scooping~\cite{Schenck2017ScoopingLearning} and excavation~\cite{Dadhich2016ExcavationSurvey} studies. These related but distinct areas operate at different scales: scooping typically involves smaller volumes and more precise control, while excavation deals with larger-scale earth moving. The task proposed by Schenck et al. focuses on manipulating a granular terrain to a certain shape~\cite{Schenck2017ScoopingLearning} by learning a predictive function of terrain shape change given an action. An optimization-based method is proposed by Yang et al. to generate excavation trajectories to excavate desired volumes of soil based on the intersection volume between the digging bucket swept volume and the terrain~\cite{yang2021optimization}. Dadhich et al. propose to use imitation-learning for rock excavation by wheel loaders, given expert demonstrations~\cite{Dadhich2016}. All of these past works are developed on a single type of material. In contrast to these methods, our work directly addresses the large deployment gaps that are likely to be found in extraterrestrial terrain sampling. 

Leveraging deep learning and its ability to process raw high-dimensional inputs such as images has become increasingly popular in the robotics domain~\cite{DRL-survey,DL-robotics-survey}. However, the performance of neural networks can deteriorate significantly when there is a deployment gap~\cite{candela2009dataset}, and for deep models to be deployed in real-world robotics tasks where deployment gaps are frequent and inevitable, they have to be adaptive. Simply fine-tuning neural networks based on online data observed during deployment to combat deployment gaps~\cite{zhuang2020comprehensive} is a possible approach for adaptation but high-capacity models tend to overfit on sparse data. A promising class of methods to train adaptive deep models is few-shot meta-learning~\cite{hospedales2021meta}, which involves training deep models on a diverse offline dataset comprising multiple related tasks. This approach enables the extraction of shared information across tasks, allowing the model to rapidly adapt to novel, yet related, tasks using only a few examples. Few-shot meta-learning has been studied in low-dimensional function regression~\cite{Hochreiter2001ICANN,Ha2017ICLR,Ravi2017ICLR}, high-dimensional vision tasks\cite{Chen2019ICLR,Snell2017NEURIPS,Vinyals2016NeurIPS,Nichol2018ARXIV}, and reinforcement learning tasks~\cite{Finn2017ICML,Mandi2022ARXIV,Ballou2022ARXIV}. However, these meta-learning methods do not necessarily work well right out of the box when applied to robot manipulation problems. As we will show in the results, current state-of-the-art methods fail to adapt quickly to OOD tasks in the terrain sampling domain. Our proposed meta-training method kCMD is designed for training deep kernel GP models. This idea of using the non-parametric deep kernel GP model for few-shot learning meta-learning has been explored before~\cite{Patacchiola2020DKT,Fortuin2019}, where kernels are trained to maximize the data likelihood on the training tasks. These methods have demonstrated robustness against overfitting when sparse online data is given. Different from these methods, kCMD explicitly trains the kernel to perform well on out-of-distribution (OOD) tasks, and as we will show in the experiments, improves the performance for few-shot scooping on novel terrains.

Once the model is trained, we use it in the Bayesian Optimization (BO) framework for decision-making. BO is a popular approach for sequential optimization where the objective function is modeled with a surrogate probabilistic model, and the action is selected in each iteration by maximizing some acquisition function that balances exploitation and exploration.   Using GP as the surrogate model for the objective function is common practice in BO. Meta-learning GP in the context of BO has also been explored before, for both GP~\cite{Huang2021BOGPMeta,Wang2018MetaBO} and deep mean and kernels~\cite{Rothfuss2021PACOH}. Closest to our approach is the work that meta-learns deep kernels and means for use in BO~\cite{Rothfuss2021PACOH}. Compared to this work, where there are dozens to hundreds of online samples, our work focuses on the few-shot regime. In addition, while this work optimizes a meta-objective for the task distribution that is computationally intractable for high-dimensional inputs, our work deals with real-world high-dimensional inputs and challenging testing scenarios that are drastically different from training scenarios.

\section*{Results}
The model in our scooping system takes as input a parametrized scooping action and a local RGB-D image patch of the terrain at the scooping location aligned with the scooping direction and predicts the mean and variance of the scooped volume. The model architecture consists of a deep mean function and a deep kernel, preceded by a shared feature extractor. The residual and variance of the volume predicted by the deep kernel are summed with the deep mean prediction to give the mean and variance of the volume. The scooping action consists of 5 parameters: the $x$, $y$ scooping location, the scooping yaw angle, the scooping depth, and a binary variable indicating whether the stiffness of the robot impedance controller is high or low. Then the model is used by a Bayesian optimization decision-maker that chooses an action from the action set (a uniform grid over action parameters) that would maximize an acquisition function that balances the scoop volume prediction and its uncertainty. Once an action is selected, an impedance controller tracks a reference trajectory generated from the action parameters.  

Our model is trained by executing a total of 5,100 random scoops on 51 terrains with different compositions and materials on the UIUC testbed, where the materials and compositions are shown in Fig.~\ref{fig:materials}. Each terrain has a unique combination of one or more {\em materials} used and their {\em composition}. Eight materials, Sand, Pebbles, Slates, Gravel, Paper Balls, Corn, Shredded Cardboard, and Mulch, are composed in three different ways to form the training data, including Single, Mixture, and Partition. The materials are placed manually in a scooping tray that is approximately 0.9\,m x 0.6\,m x 0.2\,m with varying surface features such as slopes and ridges. Some terrain examples are shown in Fig.~\ref{fig:materials}(C).

\subsection*{Testing on the UIUC Testbed}
\begin{figure}[h!]
\centering
    \includegraphics[trim=0cm 1.5cm 8cm 0cm,clip,width=1.0\linewidth]{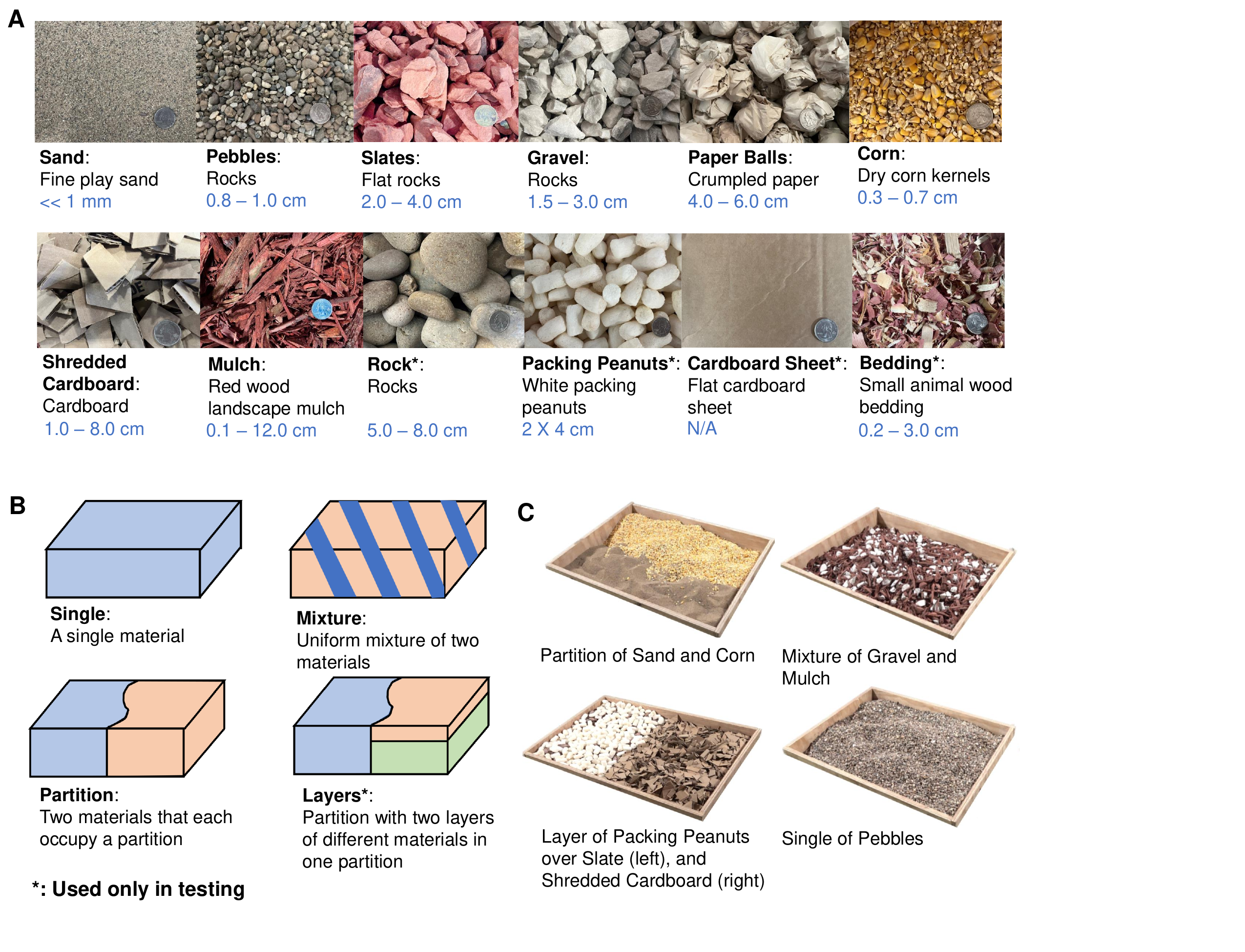}
    \caption{\textbf{All training and testing materials and compositions along with example terrains illustrating different compositions, materials, and topography, on the UIUC testbed.} Note that the Partition composition might not necessarily be half/half splits. Blue labels indicate approximate grain sizes where applicable. US quarter coin provided for scale.}
    \label{fig:materials}
\end{figure}

We first evaluate our method on the UIUC testbed, where there are 16 test terrains that contain out-of-distribution materials and compositions. We introduce 4 novel materials, which are Rock, Packing Peanuts, Cardboard Sheet, and Bedding, and a new Layers composition, described in Fig.~\ref{fig:materials}. On terrains with the Layers composition, observations do not directly reflect the composition of the terrain, and online experience is needed to infer it. Note that the Cardboard Sheet material is not scoopable. For each of the Single, Partition, Mixture, and Layers compositions, we consider 4 terrains, resulting in 16 test terrains. The 4 Single terrains are created with each of the 4 new testing materials. Material combinations on terrains with the Mixture, Partition, and Layers compositions are randomly generated from all materials but with the constraints that 1) each of the 4 novel materials is selected at least once; 2) each terrain contains at least 1 novel material. We exclude Cardboard Sheet from Mixture since it is physically impossible to create. 

\paragraph{}

\paragraph{Simulated Experiments}
We first perform 2 types of simulated experiments, \textit{simulated deployment} and \textit{prediction accuracy}, on a static test database to evaluate the performance of our methods against the state of the art. The test database consists of 100 randomly chosen scoops on each of the 16 testing terrains.

For simulated deployment, we evaluate how the model's prediction accuracy impacts adaptive decision-making performance. In this experiment, we implement a policy that only selects from the 100 actions in the dataset for the given test terrain, and the robot receives the corresponding reward observed in the dataset. A trial begins by observing a single RGB-D image as input, and the agent executes the policy until the sample reward is above a threshold $B$.  $B$ is customized for a given terrain and is defined as the 5th largest reward in that terrain's dataset in the test database. The Single Cardboard Sheet terrain is excluded in these experiments because it is not scoopable.

For prediction accuracy, we are evaluating how well each model predicts scoop volume in the $k$-shot setting. For each testing terrain, the dataset of 100 samples is first randomly split into a query set of 80 samples. Then the support set with $k$ shots is randomly drawn from the remaining 20 samples. The model prediction accuracy in terms of mean absolute error (MAE) on the query set when conditioned on the support set is evaluated.

We compare our method against three state-of-the-art meta-learning methods and one non-adaptive supervised learning (SL) baseline:
\begin{enumerate}
    \item The SL baseline uses the same network architecture as ours except that it does not contain the deep kernel. It is trained on all training data with supervised learning, and does not adapt during online testing.
    \item The first meta-learning method is implicit model-agnostic meta-learning (iMAML)~\cite{rajeswaran2019metalearning}, which is a variant that improves over the MAML~\cite{Finn2017ICML} algorithm. It is a gradient-based meta-learning algorithm that optimizes the initial weights of a neural network such that they quickly adapt to the training tasks in a few gradient descent steps. 
    \item  The second method is DKMT~\cite{Fortuin2019}, which is a meta-training method for deep kernel GPs that meta-trains the mean and kernel jointly by minimizing aggregated negative log marginal likelihood loss on all training tasks. We use the same network architecture as ours. 
    \item The third method is conditional neural processes (CNP)~\cite{garnelo2018CNP}, which is a non-kernel-based approach that learns a task representation using the support set and conditions the prediction on the query set on the learned task representation. 
\end{enumerate}
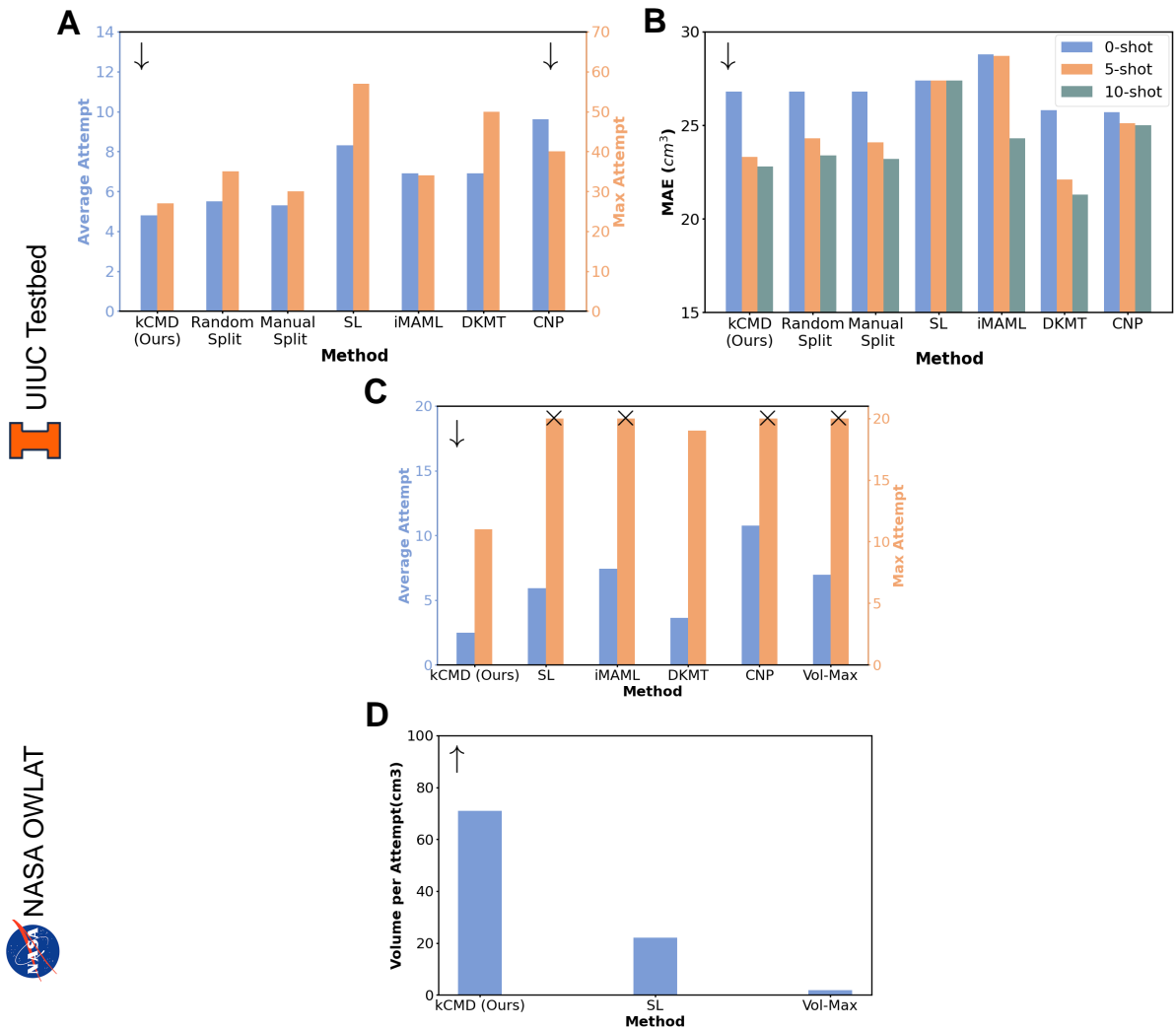
\begin{figure}[!]
\centering
\vspace{-20px}
    \includegraphics[trim=0cm 13.5cm 9cm 0cm,clip,width=1\linewidth]{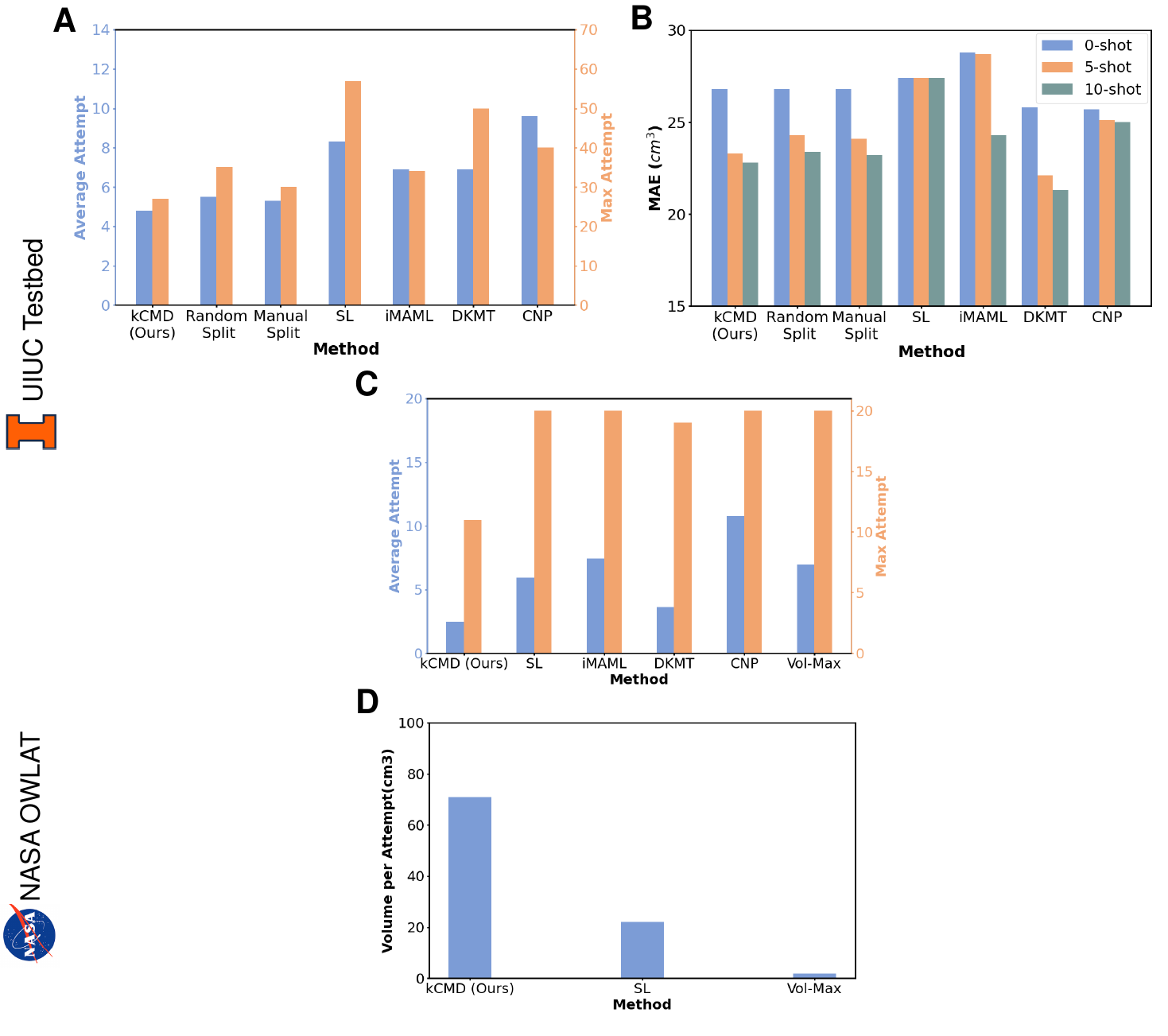}
    \put(-405,380){$\downarrow$}
    \put(-248,380){$\downarrow$}
    \put(-180,380){$\downarrow$}
    \put(-284,235){$\downarrow$}
    \put(-248.5,241){$\times$}
    \put(-221,241){$\times$}
    \put(-166.5,241){$\times$}
    \put(-139.2,241){$\times$}
    \put(-284,110){$\uparrow$}
    \caption{\textbf{Quantitative results for simulated and physical experiments on both the UIUC testbed and OWLAT.} (\textbf{A}) Simulated rollout results, where the average and max attempts to achieve success volume threshold are reported. (\textbf{B}) Simulated prediction accuracy MAE results for different shots on all testing terrains. (\textbf{C}) Physical rollout results. The allowed number of attempts is capped at 20 in order to control the experiment time. Experiments that failed at 20 attempts are denoted with $\times$. (\textbf{D}) On OWLAT, the average volume was collected for different methods with 5 attempts. For all experiments, the average across models trained with three random seeds is reported.}
    \label{fig:quantitative}
\end{figure}

In addition, we evaluate the effectiveness of leveraging OT for splitting to create maximal deployment gaps. We compare against two other ways of splitting: Random Split and Manual Split. In Random Split, the kernel and mean splits are created randomly. In Manual Split, the splitting process is created manually based on the knowledge of the underlying materials. The splits are created such that their terrain materials are different.

The results are summarized in Fig.~\ref{fig:quantitative}(A) and (B). Each model is trained 3 times with different random seeds and average results across all tasks aggregated over 3 random seeds are reported. For simulated deployment, the adaptive methods are used by a UCB decision maker with $\gamma = 2$, while the SL baseline sorts actions by the reward predicted by the mean model and greedily proceeds down the list. For prediction accuracy, the mean absolute error (MAE) is reported with 0, 5, and 10 shots. We find that kCMD significantly outperforms all baselines for simulated deployment, in terms of both average and maximum attempts used. For prediction accuracy, DKMT has the best performance while kCMD comes to a close second, both significantly outperforming all the other baselines. While the kCMD and DKMT have similar MAE reduction from 0-shot to 10-shot on average for the prediction accuracy task, DKMT performs a lot worse on the simulated deployment task. We find that this is because DKMT exhibits a high variance, even \textit{degrading} significantly in performance for some terrains from 0-shot to 10-shot adaptation. On the Single Rocks testing terrain where DKMT suffers the largest degradation, BO with the DKMT model takes as many as 44 attempts to reach the threshold for one of the random seeds. This performance degradation is due to incorrect correlations between low-quality support set samples and samples that are potentially of high quality on novel materials. Compared to using OT for splitting, Random Split and Manual Split are worse in terms of both simulated deployment and prediction accuracy. This highlights the importance of ensuring that the simulated deployment gap is maximal during training. 

\paragraph{Physical Experiments on the UIUC testbed}
We evaluate the real-world performance of our method in physical deployments on the UIUC testbed. Here, the robot executes the scooping sequence as determined from the action set by the optimizer, and each action introduces terrain shifting for the subsequent action, so the RGB-D image is re-captured after every scoop. The action set is a uniform grid over the action parameters, with 15 $x$ positions (3\,cm grid size), 12 $y$ positions (2\,cm grid size), 8 yaw angles, 4 scooping depths, and 2 stiffness settings, totaling 11520 actions. Policies are deployed on the same 15 testing terrains as the simulated deployment experiments and with the same termination threshold $B$. For each trial, a budget of 20 attempts is enforced, beyond which the trial is considered a failure. If robot trajectory planning fails for a scooping action, the next action that has the highest score according to the decision maker is selected until planning succeeds. This is done for both our method and all the baseline methods.

In addition to the baselines we compared against in the simulated experiments, we additionally employ a heuristic volume-maximizing (Vol-Max) policy, where the action is chosen to maximize the intersection between the scoop's swept volume and the terrain following a strategy proposed recently in the excavation literature~\cite{yang2021optimization}. We note that Vol-Max also does not adapt. Our method and the baseline approaches use a UCB decision maker with $\gamma = 2$, except for Vol-Max and SL that use a greedy decision maker. 

\begin{figure}[]
\centering
    \vspace{-43px}
    \includegraphics[trim=0cm 3cm 0.5cm 0cm,clip,width=0.95\linewidth, height=16cm]{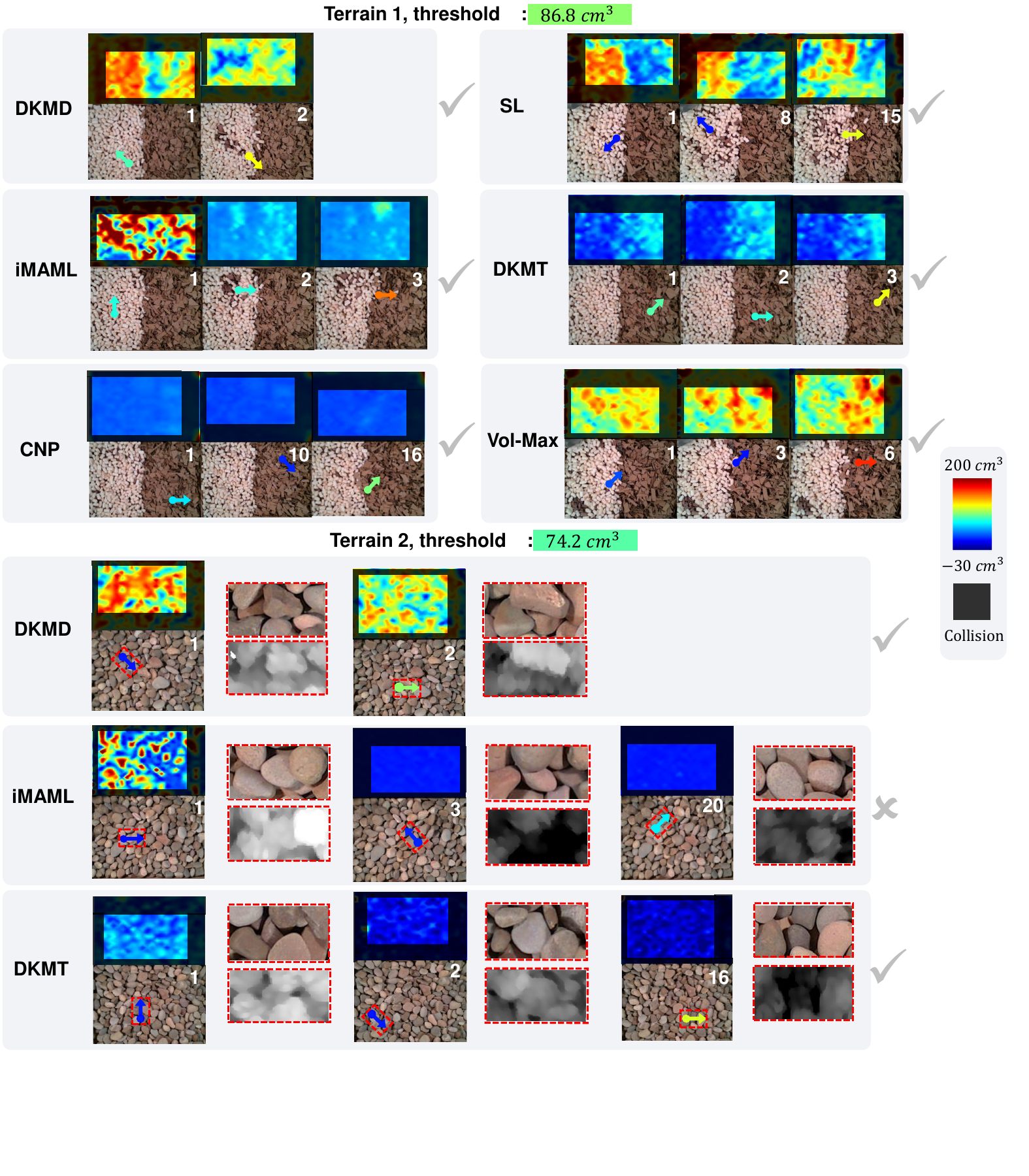}
    \put(-217,446.25){\fontsize{10}{12}\selectfont $B$}
    \put(-217,220){\fontsize{10}{12} $B$}
    \caption{\textbf{Example physical trials comparing our method and baselines.} Terrain 1 is Layers with Packing Peanuts over Slates on the left and Shredded Cardboard on the right. Terrain 2 is Single with Rock. The predicted scores by each model are visualized, with the regions that would result in the robot colliding with the terrain tray masked. The action taken and the resulting volume are shown with arrows. The volume threshold $B$ and trial success are also labeled. For each trial, the first, final, and some intermediate (if they exist) attempts are visualized, with the attempt number shown in the top right corner. For terrain 2, RGB-D patches are also visualized for more details (patches are oriented along the scooping direction, with the left edge corresponding to the edge near scoop's starting location). kCMD and iMAML are able to quickly adapt to scoop the more scoopable shredded cardboard on terrain 1, while SL and CNP struggle. On terrain 2, iMAML and DKMT, however, correlate samples incorrectly and predict low scores for promising locations very quickly, where the ideal location allows the scoop to stick into a gap between rock pieces to avoid jamming, and contains a big piece of rock in the direction of the scoop motion.}
    \label{fig:physical_rollout}
\end{figure}

Each method is run on each terrain three times. Each method except for Vol-Max is tested with three models trained with different random seeds, while Vol-Max, being a heuristic method without a learned model, is simply tested 3 times. When deploying the policies on a testing terrain, the terrain is manually reset at the start of each deployment so that surface features are consistent across trials. Note that slight terrain variations are introduced naturally during the reset. 

The average and maximal number of attempts before termination are reported in Fig.~\ref{fig:quantitative}(C). Our method outperforms the other baselines significantly. We show two representative trials on two terrains for each of the three methods in Fig.~\ref{fig:physical_rollout}. The first terrain is Layers with Packing Peanuts over Slates (left) and Shredded Cardboard (right). The Packing Peanuts material is unseen during training, and the deep mean function of kCMD predicts higher volumes on Packing Peanuts, but due to the layer of Slates underneath the scoop jams easily. After one failed scoop, kCMD quickly adapts and predicts low volumes for Packing Peanuts, and selects a scoop directed towards shredded cardboard that results in large volumes. SL takes many samples on Packing Peanuts, but eventually stops because the Slates become exposed, and Slates are in the training database and predicted to yield low volume. iMAML is also able to quickly adapt on this terrain. The mean function for DKMT happens to predict low volumes for the novel Packing Peanuts and succeeds within a few attempts. Because Shredded Cardboard has more prominent terrain features, resulting in large intersection volumes, Vol-Max always selects to scoop on Shredded Cardboard but takes 6 attempts to obtain high volumes because Vol-Max ignores the arrangement of granular particles, which has a substantial effect on the scoop outcome. The other terrain is Single Rocks. We visualize only the trials for kCMD and the two strong baselines iMAML and DKMT due to restrictions in space. While kCMD obtains a large volume quickly, iMAML and DKMT adapt the scores incorrectly, lowering scores for potentially good actions very quickly, after merely one or two actions in the support set. The ideal location on this terrain allows the scoop to stick into a gap between rock pieces to avoid jamming and contains a big piece of rock in the direction of the scoop motion that can be scooped. 

\begin{figure}[h!]
\centering
    \includegraphics[trim=0cm 17.2cm 4.7cm 0cm,clip,width=1.0\linewidth]{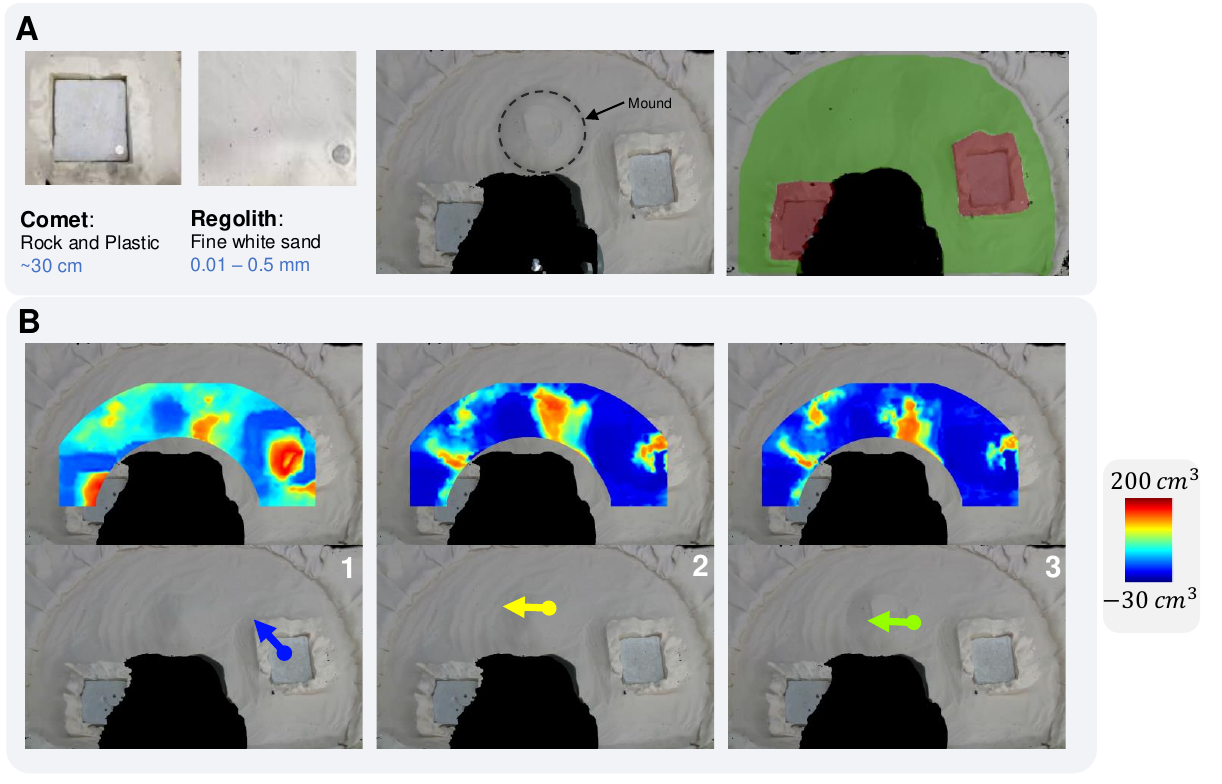}
    \caption{\textbf{Overview of materials, terrain composition, and adaptation results on the OWLAT testbed.} (\textbf{A}) The two testing materials used: scoopable Regolith (green) and unscoopable Comet (red), along with approximate grain sizes (labeled in blue) and their composition in the OWLAT testbed. (\textbf{B}) Predicted scores and chosen actions by kCMD during a rollout on this terrain for the first three attempts. Only scores within the operational workspace of the scooping arm are visualized. Action taken is shown with an arrow where the arrow's color encodes the resulting volume. After attempting a scoop on the Comet material, which yields no volume, kCMD rapidly adapts its estimates.  It then consistently scoops the Regolith material in subsequent attempts.}
    \label{fig:OWLAT_materials}
\end{figure}

\paragraph{Physical Experiments on OWLAT}
We apply our trained model directly on NASA OWLAT testbed, without any fine-tuning. This deployment was made possible because the action definition is independent of the robot arm kinematics and the scoop used on the UIUC testbed is replicated on OWLAT. OWLAT has one terrain simulant consisting of Comet and Regolith, illustrated in Fig.~\ref{fig:OWLAT_materials}. Comet is an unscoopable composition of grey comet simulant material~\cite{carey2017development} surrounded by 3D printed PLA features with rugged terrain features from a 3D scan of Devil’s Golf Course in Death Valley National Park, painted to match the Regolith’s color. Regolith is a fine sand-like material that is visually distinct from the sand used in training. The two materials are composed together to create a hypothetical representation of the ocean world terrain. The scoopable Regolith region is designed with mounds that have heights comparable to those of the unscoopable Comet regions. 

Our UIUC testbed experiments demonstrated kCMD's superior performance among adaptive methods. To evaluate its effectiveness in OWLAT, we compare kCMD with SL and Vol-Max, the two non-adaptive baselines that represent approaches in the excavation domain~\cite{Lu2021Excavation,yang2021optimization}, and report the average volumes collected per attempt with a fixed budget of 5 attempts, emulating rover missions~\cite{NASA2017}. The quantitative results are shown in Fig.~\ref{fig:quantitative} (D). Due to the large prominent features of the Comet, Vol-Max selects to scoop at the Comet region and obtains low volumes. SL starts with the Comet region and fails to modify its policy in response to the data observed online, continuing the ineffective scooping attempts in the Comet region, akin to Vol-Max. kCMD initially targeted the Comet region but quickly adapted to scoop at the Regolith, and we show a representative trial on the OWLAT testbed with kCMD in Fig.~\ref{fig:OWLAT_materials}.

\section*{Discussion}
This paper introduced a novel method for granular material manipulation under domain shift that uses a vision-based
few-shot learning approach to adapt quickly
to small amounts of online data. Our novel meta-training procedure, Deep Kernel Calibration with Maximal Deployment Gaps, demonstrates encouraging results for meta-training generalizable and adaptive high-capacity models.

Although our method performed well in the observed experiments, a deployed robot may encounter exotic materials in which correlations between appearance, action, and result are drastically different than those observed in training. In such cases, the learned model may mislead the robot to perform poorly, and possibly even worse than exhaustive uninformed sampling. A potential remedy for this problem would be to adapt the GP kernel online. We leave investigating this approach to future work. 

Our current method relies on impedance control for reactive movement to track a target trajectory, and it frequently jams in challenging rocky terrains. A possible area of future work is utilizing the visual appearance of particle movements and the contact force experienced when executing a scoop action, which could be very informative about the underlying terrain to improve performance. We hope to adopt terrain-adaptive feedback controllers that alter the movement strategy during a scoop to further improve sample volumes.

Finally, we would like to explore more complex rewards other than sampling volume, such as the outcome of a scientific assay from a sample analysis instrument. 

\section*{Method and Materials}
In this section, we first detail the problem formulation and the scooping setups, and then describe our system and method. 

\paragraph{Problem Formulation}
We formulate the scooping problem as a sequential decision-making task, where the robot in each \textit{episode} observes the terrain RGB-D image $o \in \mathcal{O}$ and uses a \textit{scooping policy} to apply $a \in \mathcal{A}(o)$ where $\mathcal{A}(o)$ is a discrete set of parameterized, observation-dependent scooping motions. The reward $r \in \mathcal{R} $ of a scoop is the scooped volume. Throughout the paper, a terrain is defined as a unique {\em composition} of one or more {\em materials}, where a material is composed of particles with consistent geometry and physical properties.
\newline

Presented with a target terrain $T_*$, the robot's goal is to find a scoop whose reward is above a threshold $B$. In planetary missions, for example, $B$ could be the minimal volume of materials needed to perform an analysis. During the $n$-th episode, the robot knows the history of scoops on this terrain $H = \{(o^j,a^j,r^j)\,|\,j=1,\ldots,n-1\}$, which we also refer to as the \textit{online support set}. Note that the support set only contains samples of low quality, i.e. below $B$, because otherwise the goal would already have been achieved. 
\newline

The robot has access to an \textit{offline} prior scooping experience, which consists of a set of $M$ terrains $\{T_1, \dots, T_M \}$, and a training dataset $D_i = \{(o^j,a^j,r^j)\,|\,j=1,\ldots,N_i\}$ of past scoops and their rewards for each terrain $i=1,...,M$.

For a terrain, we suppose a \textit{latent variable} $\alpha$ characterizes its composition, material properties, and topography, which are only indirectly observed. Let $\alpha_*$ characterize $T_*$ and $\alpha_i$ characterize $T_i$ for $i=1,\ldots,M$. Moreover, the observation is dependent on the latent variable, and an action's reward $r\equiv r(\alpha,a)$ is also an unknown function of the action and latent variable. Standard supervised learning applied to model $r\approx f(o,a)$ will work well when $\alpha_*$ is within the distribution of training terrains, and $\alpha_*$ is uniquely determined by the observation $o$ or the reward is not strongly related to  unobservable latent characteristics. However, when $T_*$ is out of distribution or the observation $o$ leaves ambiguity about latent aspects of the terrain that affect the reward, the performance of the learned model will degrade. 

Considering the limitations of supervised learning in this setting, online learning from $H$ has the potential to help the robot perform better on $T_*$.  Meta-learning attempts to model the dependence of the reward or optimal policy on $\alpha$, either with explicit representations of $\alpha$ (e.g., conditional neural processes~\cite{garnelo2018CNP}) or implicit ones (e.g., kernel methods~\cite{Patacchiola2020DKT}, which are used here).

\paragraph{Scooping Setups}
We first train and test our method on a university (UIUC) testbed, then directly deploy the model on the NASA Ocean Worlds Lander Autonomy Testbed (OWLAT)~\cite{nayar2021development} without finetuning. The two setups are shown in Fig.~\ref{fig:platforms}. The UIUC testbed includes a UR5e arm with a scoop mounted on the end-effector, an overhead Intel RealSense L515 RGB-D camera, and a scooping tray that is approximately 0.9\,m x 0.6\,m x 0.2\,m. OWLAT is a high-fidelity testbed developed to validate autonomy algorithms for future ocean world missions. It serves as a state-of-the-art platform for simulating various potential future planetary missions over a wide range of dynamic environments, including surface operations on small
bodies where recreating the dynamics in low gravity is critical. The testbed hardware consists of a 7-DOF Barrett WAM7 robotic arm with a host of interchangeable end-effector tools including the manipulator, an Intel Realsense D415 mounted on a pan-tilt mount for 3D perception, and force-torque sensors located at the interface between the arm and the platform and also at the end of the arm’s wrist. 

We consider a variety of materials and compositions for the offline database collected on the UIUC testbed. The materials used in this project are listed in Fig.~\ref{fig:materials}. The offline database contains materials Sand, Pebbles, Slates, Gravel, Paper Balls, Corn, Shredded Cardboard, and Mulch. The testing terrains on the UIUC testbed also include Rock, Packing Peanuts, Cardboard Sheet and Bedding, which significantly differ from the offline materials in terms of appearance, geometry, density, and surface properties. The terrain compositions used are listed in Fig.~\ref{fig:materials}. The offline database contains the Single, Mixture, and Partition compositions, while the testing set also contains the Layers composition. On terrains with the Layers composition, observations do not directly reflect the composition of the terrain, and online experience is needed to infer it. All terrains are constructed manually, with varying surface features (e.g. slopes, ridges, etc.) with a maximum elevation of about 0.2\,m and a maximum slope of 30$^{\circ}$. Some terrain examples are demonstrated in Fig.~\ref{fig:materials}. We also observe that the scooping outcomes show high variance because many terrain properties are not directly observable, such as the arrangement and geometry of the particles beneath the surface. For evaluation on OWLAT, we used a terrain designed by subject matter experts using out-of-distribution materials, Comet and Regolith, which is detailed in the Results section. 

\begin{figure}[]
\centering
    \includegraphics[trim=0cm 6cm 1cm 0cm,clip,width=1\linewidth]{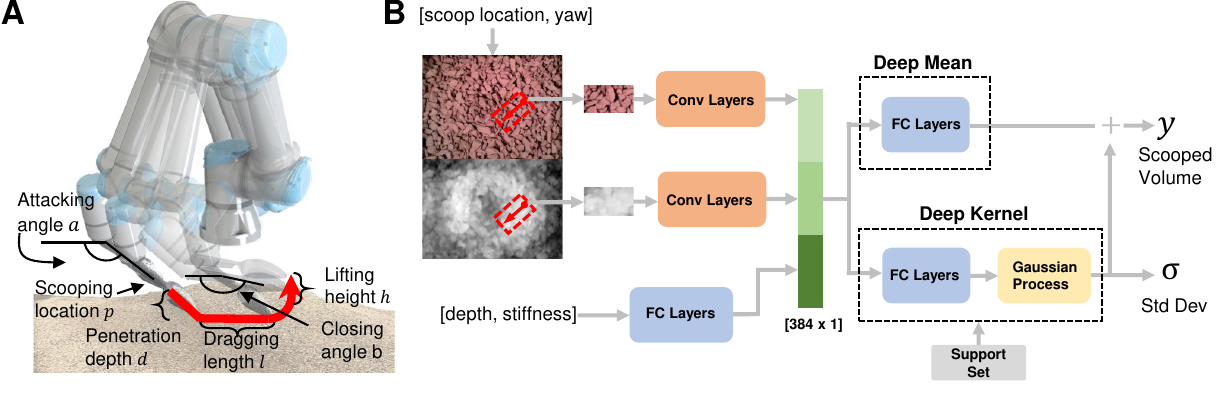}
    \caption{\textbf{The reference scooping trajectory definition and model architecture.} \textbf{(A)} The parameterized reference trajectory definition. \textbf{(B)} RGB-D image patches are cropped given the scoop location and the yaw angle of the scoop direction, and are fed into the neural network. Deep kernel and the deep mean share a common feature extractor. ``FC" denotes fully-connected layers and ``Conv" denotes convolution layers. The radial basis function kernel is used for the GP.}
    \label{fig:architecture}
\end{figure}

A \textit{scoop action} is a parameterized trajectory for a scoop end effector that is tracked by an impedance controller. We follow the common practice in the excavation literature~\cite{Lu2021Excavation,sing1995synthesis} to define a scooping trajectory, shown in Fig.~\ref{fig:architecture}(A), where the scoop has a roll angle of 0 and stays in a plane throughout the trajectory. The scoop starts the trajectory at a location $p$, penetrates the substrate at the attack angle $a$ to a penetration depth of $d$, drags the scoop in a straight line for length $l$ to collect material, closes the scoop to an angle $\beta$, and lifts the scoop with a lifting height $h$. We assume that the scoop always starts scooping at the terrain surface, which can be determined from the depth image. The impedance controller of the end-effector is configured with stiffness parameters $b$.

To reduce the action space, we manually tuned the parameters that have minimal impact on the scooping outcome, fixing the attack angle $a$ at 135$^{\circ}$, the dragging length $l$ at 0.06\,m, the closing angle $\beta$ at 190$^{\circ}$, and the lifting height $h$ at 0.02\,m. In addition, we set two options for the impedance controller stiffness $b$, corresponding to soft and hard stiffness, where the linear spring constants are 250\,N/m and 750\,N/m and the torsion spring constants are 6\,Nm/rad and 20\,Nm/rad, respectively. Therefore, the action is specified by the starting $x$, $y$ position and yaw angle of the scoop, the scooping depth $d$, and stiffness $b$.

To measure the scooped volume, the scoop is moved to a fixed known pose, after which a height map within the perimeter of the scoop is obtained from the depth image. The volume is then calculated by integrating the difference between this height map and the height map of an empty scoop at the same pose collected beforehand.

The offline database contains data on 51 terrains, all with unique combinations of materials and compositions. Out of these terrains, 8 are Single, 25 are Partition, and 18 are Mixture. The materials used are randomly selected from the training materials. For each terrain, we collect 100 random scoops, sampled uniformly with random $x$, $y$ positions in the terrain tray, random yaw angle from a set of 8 discretized yaw angles, 45$^\circ$ apart, random depth in the range of 0.03\,m to 0.08\,m, and random stiffness (either ``hard'' or ``soft''). Sometimes trajectory planning of the robot manipulator for a sampled scoop can fail due to kinematic constraints. If so, the scoop action is discarded and sampling continues until planning is successful. The average scooped volume across the offline database is 31.3\,cm$^3$, and the maximum volume is 260.8\,cm$^3$. 

\paragraph{Deep Kernel Gaussian Process Model}
Our approach models the reward's dependence on the observation $o$, action $a$, and history $H$ as a deep kernel Gaussian process (GP) model.  Our Deep Kernel Calibration with Maximal Deployment Gaps (kCMD) method meta-trains the deep kernel to perform well under simulated deployment gaps extracted from the training set.  With such a model, the predicted reward and its variance are used at each step to optimize the chosen action using Bayesian optimization. We will first describe our proposed deep GP model, and then describe training the model with kCMD. For convenience of notation, we let $x=(o,a)$ denote an observation-action pair, and $y=r$ denote a reward. Let us also separate the training datasets into sequences of dependent variables $D_i^y = \{y_i^j\,|\,j=1,\ldots,N_i\}$ and independent variables $D^x_i = \{x_i^j\,|\,j=1,\ldots,N_i\}$, where $i$ denotes the $i$-th task dataset. 

A GP models function $f$ as a collection of random variables $f(x)$ which are jointly Gaussian when evaluated at locations $x$~\cite{GPML}. A GP is fully specified by its mean function $m(\cdot)$ and kernel $k(\cdot,\cdot)$, which is the covariance function:
\begin{equation}
    f(x) \sim \mathcal{GP}(m(x),k(x,x')).
\end{equation}

Given $n$ existing observed function values $\mathbf{y} = [y_1,\dots,y_n]^T$ at $\mathbf{x} = [x_1,\dots,x_n]^T$, GP regression predicts the function values at new point $x^*$ as a Gaussian distribution:
\begin{equation}
    P(y^*|\mathbf{x}, \mathbf{y}, x^*) \sim 
    \mathcal{N}(m(x^*) + \mathbf{k}\mathbf{K}^{-1}\bar{\mathbf{y}},k(x^*,x^*) - \mathbf{k}\mathbf{K}^{-1}\mathbf{k}^T ).
\end{equation}
Here,
\begin{equation*}
\begin{split}
        &\mathbf{K} = \begin{bmatrix}
                k(x_1,x_1) & \cdots & k(x_1,x_n)\\
                \vdots & \ddots & \vdots\\
                k(x_n,x_1) & \cdots & k(x_n,x_n)\\
                \end{bmatrix} + \sigma_n^2\mathbf{I},  \\
        &\mathbf{k} = \begin{bmatrix}
            k(x^*,x_1), & \cdots, & k(x^*,x_n)\\
            \end{bmatrix},  \\
        &\bar{\mathbf{y}} = [y_1 - m(x^*),\dots,y_n - m(x^*)]^T,
\end{split}
\end{equation*}
where $\sigma_n$ is the standard deviation of noise at an observation and $\bar{\mathbf{y}}$ is the residual. A typical choice for the mean function is a constant mean and the radial basis function kernel (RBF) is a popular kernel of choice~\cite{GPML}. The mean constant, kernel function parameters, and $\sigma_n$ can be hand-picked if there is prior knowledge of $f$. In practice, such knowledge is usually not available and they are estimated from data with type-II maximum likelihood by minimizing the negative log marginal likelihood (NLML):
\begin{equation}
\begin{split}
    -\log P(&
    \mathbf{y}|\mathbf{x}, \theta) = \frac{1}{2} \log |\mathbf{K} + \sigma_n^2 \mathbf{I}| \\
    &+ \frac{1}{2}(\mathbf{y} - m(\mathbf{x}))^T(\mathbf{K} + \sigma_n^2 \mathbf{I})(\mathbf{y} - m(\mathbf{x})) \\
    &+ c,
    \end{split}
\label{eq:LML}
\end{equation}
where $\theta$ denotes all the parameters to be determined and $c$ is a constant.

Deep kernels leverage neural networks to improve the scalability and expressiveness of kernels~\cite{wilson2016DKL} and have been proposed for the few-shot setting~\cite{Patacchiola2020DKT}. For deep kernels, an input vector is mapped to a latent vector using a neural network before going into the kernel function $k(g_\theta(\cdot), g_\theta(\cdot))$, where $g_\theta(\cdot)$ is a neural network with weights $\theta$. Additionally, deep kernels were extended to use deep mean functions $m_\theta(\cdot)$~\cite{Fortuin2019} to learn more expressive mean functions. Our proposed deep GP contains both a deep kernel and a deep mean. The model takes in the RGB-D image observation and action parameters to predict the scooped volume. Specifically, we use a localized RGB-D image patch - a small portion of the full image starting at the sampling location and aligned with the yaw angle - rather than the entire image, along with the scooping depth and binary stiffness variable action parameters as input. This image patch contains most of the information needed to evaluate a scoop, and is much more computationally efficient than processing the entire image. The model architecture is shown Fig.~\ref{fig:architecture}(B). The kernel and mean function share the same feature extractor, which is a convolutional neural network, and have separate fully connected layers. 

\paragraph{Deep Kernel Calibration with Maximal Deployment Gaps}
The neural network parameters and the kernel parameters of a deep GP can be jointly trained over the entire training set with the same NLML loss as Eqn.~\ref{eq:LML}, where $\theta$ contains the neural network parameters. However, this approach does not typically train kernels that are well-tuned to individual tasks because it aggregates the data from all tasks together. Instead, meta-training may be realized with stochastic gradient descent with each batch containing the data for a single task, i.e. minimizing the following aggregate loss:
\begin{equation}
   \min_{\theta}\sum_j\sum_i -\log P(D_{i,j}^y|D_{i,j}^x, \theta), 
   \label{eqn:NLML}
\end{equation}
 where $D_i^y$ and $D_i^x$ are the target variables and input variables of task $j$. This approach, here-in-after referred to as Deep Kernel and Mean Transfer (DKMT), has been proposed in the meta-learning literature~\cite{Fortuin2019, Rothfuss2021PACOH}.
 
 DKMT has potential problems with out-of-distribution tasks. During training, the residuals seen by the kernels are residuals of the deep mean on the training tasks, which could be very different compared to the residuals on tasks out of distribution of the training terrains. As our experiments will show, this feature leads to the kernels being poorly calibrated. Another potential issue is the over-fitting of the deep mean function. The first two terms of NLML in Eqn.~\ref{eq:LML} are often referred to as the \textit{complexity penalty} and \textit{data fit} terms, where the \textit{complexity penalty} regularizes the deep kernels~\cite{GPML}. However, there is no regularization of the deep mean function. As a result, the deep mean can potentially overfit on all the training data, so the residuals will be close to zero.  

kCMD addresses these issues by encouraging the residuals seen in kernel training to be representative of the residuals seen in out-of-distribution tasks with novel materials. The idea is to split the training terrains into a mean training set and a kernel training set, where each set contains terrains that are maximally different from each other. This is achieved by using optimal transport (OT)~\cite{torres2021survey}, which is a principled approach for calculating the distance between two probability distributions, to measure the difference between two datasets. To perform the splitting given the pairwise distance between datasets, each split is obtained by first randomly selecting a task dataset as the reference task dataset $D_r$, and splitting all task datasets into one that contains the most similar datasets to $D_r$, according to OT, and one that contains the most different. Then, the mean is trained on the mean training set to minimize error and the GP is trained on the residuals of the mean model on the kernel training set. As we showed in the results, such a simulated deployment process with maximal deployment gaps leads to trained kernels that generalize better to novel terrains. The optimal transport between two probability distributions can be calculated given samples and a corresponding cost function that computes the distance between sample points. To take into account both the observation, action, and the corresponding volume of the data samples, we consider the following cost function, similar to what is done in~\cite{alvarez2020geometric,courty2017joint}:
\begin{equation}
    d\bigl((o_i,a_i,r_i), (o_j,a_j,r_j)\bigr) = \bigl((d_{img}(o_i,o_j)/C_1)^2 + (\lVert a_i -a_j)\rVert/C_2)^2 + (\lVert r_i - r_j\rVert/C_3)^2\bigr)^{1/2}, 
\end{equation}
where $d_{img}$ measures the distance between the RGB and depth image patches and $C_1$, $C_2$, and $C_3$ are normalization constants, which are calculated as the largest norm of observation, action, and volume across the entire dataset, respectively. $d_{img}$ is calculated by first obtaining feature vectors for the RGB-D image patch pair, where each feature vector is obtained by computing the histogram of each of the RGB and depth channels with 32 bins and concatenating into a 128-dimensional feature vector. Then $d_{img}$  is the $L_2$ norm between the two feature vectors. Directly solving for the optimal transport plan between the two distributions is computationally intensive and we opt to approximate it with the Sinkhorn divergence~\cite{feydy2019interpolating}, which is calculated using an efficient implementation from the Geomloss library~\cite{feydy2019interpolating}.

One concern for splitting the training dataset is that it limits the amount of training data available for the mean and kernel. Therefore, we repeat this process similarly to $k$-fold cross-validation, in which each fold has a separate mean model trained on the mean split for that fold, and then the residuals for that model on the kernel split are used to define the kernel loss for that fold. A common kernel is trained using losses aggregated across folds. 

\begin{figure}[]
\centering
    \includegraphics[trim=0cm 2cm 7cm 0cm,clip,width=0.95\linewidth]{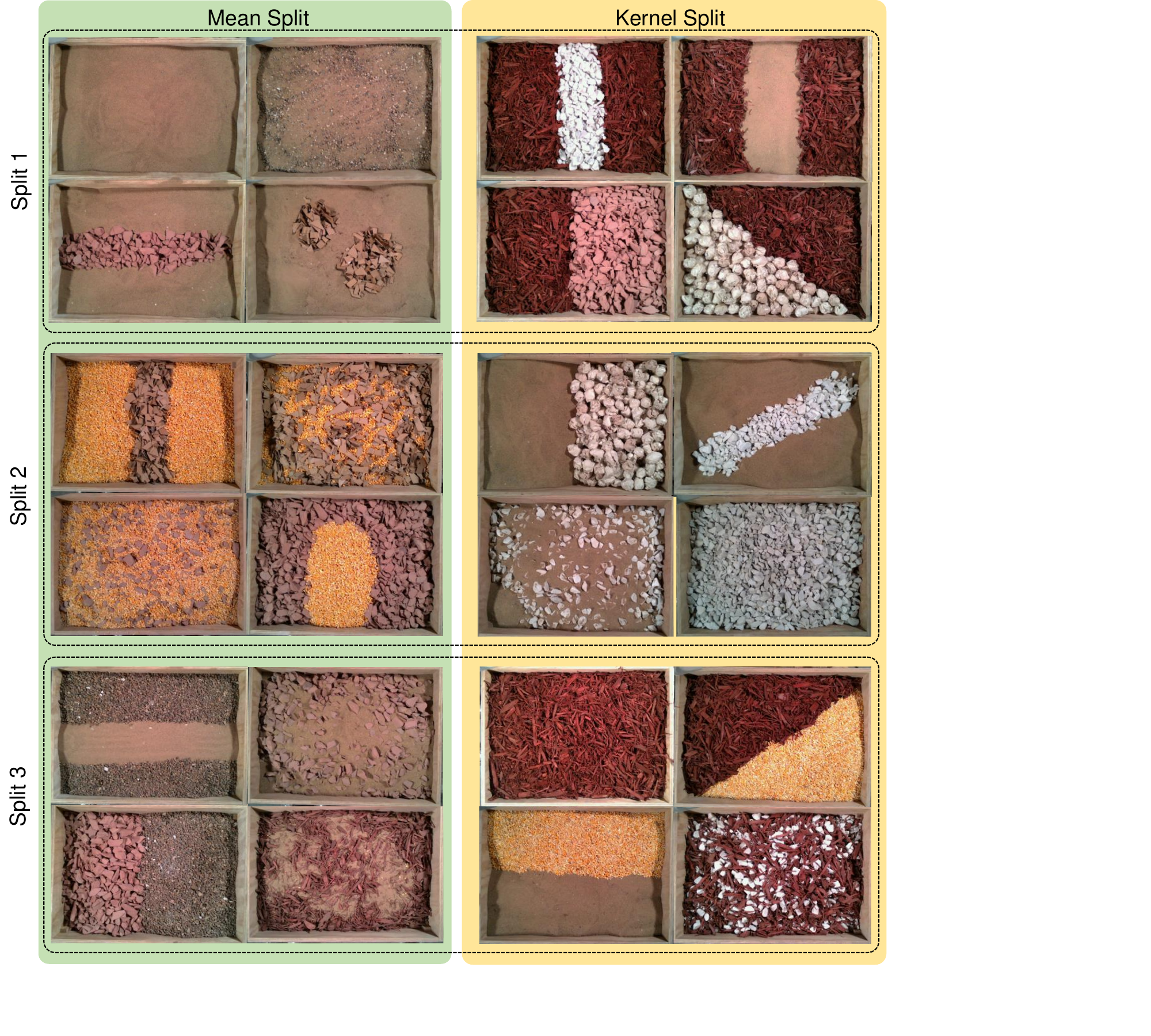}
    \caption{\textbf{kCMD split visualization.} Three of the ten splits are visualized here, where each split is denoted by the dashed outline. The top left terrain in each split is the reference task for this split during the splitting process. The three most similar terrains to the reference task are visualized in the mean split, while the four most dissimilar terrains are visualized for the kernel split. } 
    \label{fig:kCMD_split}
\end{figure}

The overall training procedure for our proposed kCMD method is detailed in Algorithm~\ref{alg:training}. The feature extractor and the deep mean ($\theta_f$ and $\theta_m$) are first jointly trained on all training tasks with standard supervised learning and saved to disk (Line 2). Subsequently, the kernel residuals are collected for each of the $K$ splits (Line 3-9). Some splits are visualized in Fig.~\ref{fig:kCMD_split}. For each mean split, the feature extractor and deep mean with weights $\theta_f^k$ and $\theta_m^k$ are trained from scratch, and the kernel splits' residuals are collected. The deep kernel parameters $\theta_k$ are then meta-trained with the NLML loss on all the collected residuals across the N splits. Finally, the $\theta_f$ and $\theta_m$ are loaded from disk to return the final model, where $\theta_f^k$ and $\theta_m^k$ are discarded (Line 10-12). Note that to encourage the kernel weights $\theta_k$ to be tuned to the feature extractor $\theta_f$, when training the feature extractor from scratch for collecting the kernel residuals during each split (Line 7) we also add an $L_2$ regularization term with a coefficient of 1 to encourage the weights $\theta_f^k$ to stay close to $\theta_f$. 

\begin{algorithm}[h]
\SetAlgoLined
\textbf{Input:} All task data $\{D_i\}$; number of folds $K$\;
Train the feature extractor and mean model with weights $\theta_f$ and $\theta_m$ on $\{D_i\}$ with standard supervised learning and save to disk\;
\tcp{Collect mean residuals on simulated deployment gaps}
Initialize empty residual data $E \leftarrow \emptyset$ \;
\For{k = 1, $\cdots$, K}{
Randomly initialize the feature extractor weights $\theta_f^k$ and deep mean weights $\theta_m^k$\;
Randomly select a reference task dataset $D_r$ and split all task datasets into one that contains the most similar dataset to $D_r$, according to OT, and one that contains the most different ones: $S^k_{mean}$ and $S^k_{kernel}$ \;
Train $\theta_f^k$ and $\theta_m^k$ using a mean squared error (MSE) loss on all the data in $S^k_{mean}$, while penalizing the difference between $\theta_f^k$ and $\theta_f$ with the $L_2$ penalty $\lVert \theta_f^k - \theta_f\rVert_2^2$ \; 
\For{$D_i \in S^k_{kernel}$}
{Collect the residuals of the mean model $\delta_i^j = r_i^j - m_{\theta^k}(x_i^j)$, $j=1,...,N_i$. Construct the inputs $\hat{D}_i^x = D_i^x$ and the outputs $\hat{D}_i^y = \{\delta_i^j\,|\,j=1,...,N_i\}$ to predict the residuals.  $E \leftarrow E \cup \hat{D}_i$\;}}
\tcp{Meta-training deep kernels}
Train deep kernel parameters $\theta_k$ with database $E$ with Eqn.~\ref{eqn:NLML}. (Note that the associated features extractor weights $\theta_f^k$ for each task need to be loaded and frozen at the start of each batch training) \;
Reload $\theta_f$ and $\theta_m$ from disk\;
 \Return  $\theta_m,\theta_f,\theta_k$\;
 \caption{Meta-training with Controlled Deployment Gaps}
 \label{alg:training}
\end{algorithm}

\paragraph{Bayesian optimization decision-maker}
To use a reward model in the scooping sequential decision-making problem, the decision-maker maximizes a score $s(o,a)$ over the action $a$: $\pi(o) = \arg \max_{\mathcal{A}(o)} s(o,a)$.  A greedy optimizer would use the mean as the score, $s(o,a) = m(o,a,H)$ where $m(o,a,H)=E[r | r \sim p(R\,|\,o,a,H)]$, but this definition does not adequately explore actions for which the prediction is uncertain. Instead, a Bayesian optimizer uses an acquisition function that also takes uncertainty into account. For example, the upper confidence bound (UCB) method defines the acquisition function $s_{UCB}(o,a)=m(o,a,H) + \gamma \cdot \sigma(o,a,H)$, where $\sigma(o,a,H)=Var[r | r \sim p(R\,|\,o,a,H)]^{1/2}$ is the standard deviation of the prediction and $\gamma > 0$ is a parameter that encourages the agent to explore actions whose outcomes are more uncertain.

\paragraph{Model Training}
We select $K = 10$ for all training. PyTorch~\cite{PyTorch} and GPyTorch~\cite{gardner2018gpytorch} are used to implement the neural networks and GP. The Adam optimizer is used for training. Learning rates of 5e-3 and 1e-2 are used for the training of the deep mean and deep kernel, respectively. These values are hand-picked by inspecting the training loss without extensive hyperparameter tuning. For training the mean, 10$\%$ of the training data is used for validation and early stopping based on the validation loss with patience of 5 is used to select the training epochs. Early stopping with a patience of 5 based on the training loss is used to select the training epochs for the deep kernel. We also apply data augmentation, where for both mean and kernel training, random vertical flips of the images are used since flipping vertically would not change the predicted volume. For mean training, random hue jitter and random depth noise are also applied. The training process takes less than 2 hours on a hardware setup consisting of an i7-9800x CPU, a 2080Ti GPU, and 64GB of RAM.

\bibliography{references.bib}

\bibliographystyle{Science}

\section*{Acknowledgments}
This work was supported by NASA Grant 80NSSC21K1030. Part of this work was carried out at the Jet Propulsion Laboratory, California Institute of Technology, under a contract with the National Aeronautics and Space Administration (80NM0018D0004).

The authors declare that they have no competing interests. All data are available in the main text or the supplementary materials.The summary of the author contributions are:
\begin{itemize}
\setlength\itemsep{-0.5em}
    \item Conceptualization: YZ, PT, MO, KH
    \item Methodology: YZ, PT, KH
    \item Software: YZ, PT, ET, AG, EK
    \item Validation: YZ, PT
\item Formal analysis: YZ, PT
\item Investigation: YZ, PT, ET, AG, EK, HD, MO, KH
\item Data curation: YZ, PT
\item  Visualization: YZ, PT
\item Funding acquisition: MO, KH
\item Supervision: HD, MO, KH
\item Writing – original draft: YZ, PT
\item  Writing – review and editing: YZ, PT, ET, AG, HD, MO, KH
\end{itemize}

\clearpage

\end{document}